\documentclass[10pt,twocolumn,letterpaper]{article}
\usepackage{authblk}
\usepackage{cvpr}              

\usepackage[dvipsnames]{xcolor}

\definecolor{cvprblue}{rgb}{0.21,0.49,0.74}
\usepackage[pagebackref,breaklinks,colorlinks,citecolor=cvprblue, pdftex]{hyperref}
\usepackage{colortbl}
\usepackage{graphicx}
\usepackage{amsmath}
\usepackage{multirow}
\usepackage{xcolor}
\usepackage{mathtools}
\usepackage{pifont}

\def\paperID{18102} 
\def\confName{CVPR}
\def\confYear{2025}

\usepackage[accsupp]{axessibility}  

\newcommand{\Fref}[1]{Fig~\ref{#1}}

\newcommand{\Sref}[1]{Sec~\ref{#1}}
\newcommand{\Tref}[1]{Table~\ref{#1}}

\newcommand{\bspline}{B-spline Mapper}
\newcommand{\fourier}{Fourier Mapper}
\newcommand{\cmark}{\ding{51}}%

\makeatletter 
\renewcommand\AB@affilsepx{\hspace{1cm} \protect\Affilfont}\makeatother

\title{BF-STVSR: B-Splines and Fourier---Best Friends for High Fidelity Spatial-Temporal Video Super-Resolution}

\author[1]{Eunjin Kim\thanks{Equal contribution}}
\newcommand\CoAuthorMark{\footnotemark[\arabic{footnote}]}
\author[1]{Hyeonjin Kim\protect\CoAuthorMark}
\author[2]{Kyong Hwan Jin}
\author[1]{Jaejun Yoo}

\affil[1]{Ulsan National Institute of Science and Technology (UNIST)}
\affil[2]{Korea University} 

\affil[ ]{\tt\small \{eunjin.kim, hyeonjin.kim, jaejun.yoo\}@unist.ac.kr, kyong\_jin@korea.ac.kr }

\begin{document}

\maketitle
\begin{abstract}

While prior methods in Continuous Spatial-Temporal Video Super-Resolution (C-STVSR) employ Implicit Neural Representation (INR) for continuous encoding, they often struggle to capture the complexity of video data, relying on simple coordinate concatenation and pre-trained optical flow networks for motion representation.  Interestingly, we find that adding position encoding, contrary to common observations, does not improve---and even degrades---performance. This issue becomes particularly pronounced when combined with pre-trained optical flow networks, which can limit the model’s flexibility. To address these issues, we propose \textbf{BF-STVSR}, a C-STVSR framework with two key modules tailored to better represent spatial and temporal characteristics of video: 1) {\bspline} for smooth temporal interpolation, and 2) {\fourier} for capturing dominant spatial frequencies. Our approach achieves state-of-the-art in various metrics, including PSNR and SSIM, showing enhanced spatial details and natural temporal consistency. Our code is available \href{https://github.com/Eunjnnn/bfstvsr}{here}.

\end{abstract}    
\section{Introduction}
\label{sec:intro}

Enhancing low-resolution, low-frame-rate videos to high-resolution, high-frame-rate quality is crucial for delivering seamless user experiences. To address this, deep learning approaches for Video Super-Resolution (VSR) \cite{caballero2017real, sajjadi2018frame, chan2021basicvsr, chan2022basicvsrpp} and Video Frame Interpolation (VFI) \cite{huang2022rife, niklaus2020softmax, park2021asymmetric, reda2022film, zhang2023extracting} have been extensively studied. VSR typically enhances spatial resolution of target frames by leveraging information from neighboring frames, while VFI improves temporal resolution by predicting inherent motion in video data. However, many existing methods are limited by fixed scaling factors determined during training, which restricts their adaptability to real-world applications. 

\begin{figure}[t]
    \centering
    \includegraphics[width=\linewidth]{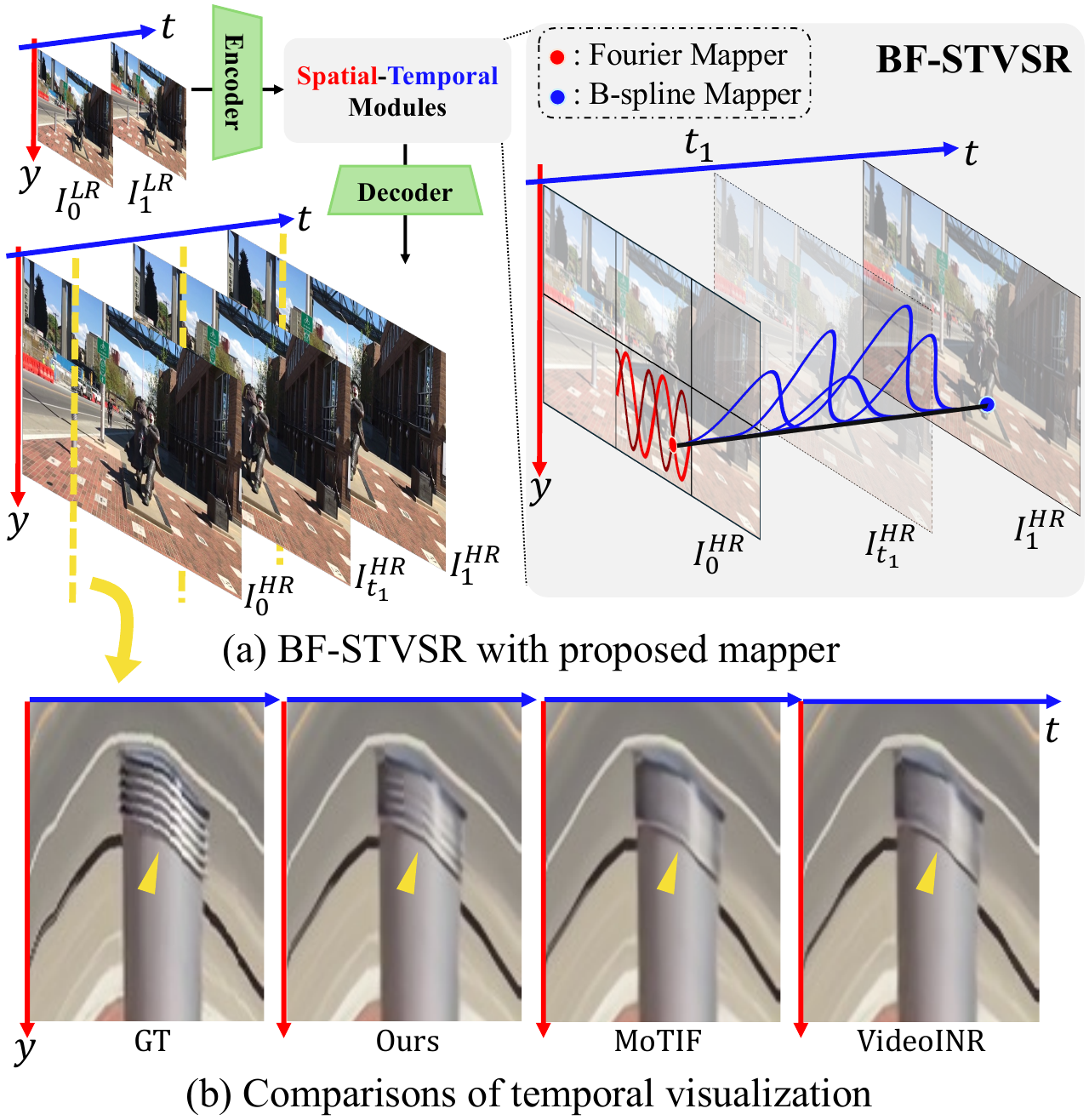}
    \label{fig1a}
    \caption{Illustration of BF-STVSR and results. (a) BF-STVSR captures the high-frequency spatial features by {\fourier} and interpolates temporal information smoothly via {\bspline}. 
    (b) We visualize the changes of the interpolated frames over time $t$ for a selected x-axis (yellow vertical line in (a)). 
    } 
\end{figure}


On the other hand, Implicit Neural Representation (INR) has recently garnered attention for its capability to represent signals continuously through a multi-layer perceptron (MLP), making it a promising approach for super-resolution (SR) tasks  \cite{liif, lte, btc, hu2019meta}. Building on these advancements, recent studies have extended INR to video data to achieve Continuous Spatial-Temporal Video Super-Resolution (C-STVSR), which enables spatial and temporal interpolation simultaneously at arbitrary scales \cite{chen2022vinr, chen2023motif}. VideoINR \cite{chen2022vinr} was the first method to map spatiotemporal coordinates ${(x,y,t)}$ to backward motion field, facilitating backward warping of spatial features to any temporal coordinate. MoTIF \cite{chen2023motif} improved on this by replacing the backward warping with forward warping, using softmax splatting \cite{niklaus2020softmax}. In addition, to facilitate the learning in an explicit way, MoTIF supply optical flow maps estimated between reference frames as contextual information, using the pre-trained optical flow network, RAFT \cite{teed2020raft}.

While VideoINR and MoTIF successfully integrate INR into the C-STVSR task, they have notable limitations. Specifically, they generate target features by encoding latent features that are simply concatenated with target coordinates, without employing advanced position encoding techniques. This simple coordinate concatenation may fall short in capturing the nuanced details of spatial and temporal features, especially for motion features, which are inherently complex and dynamic. Consequently, both models struggle to retain high-frequency information in the encoded spatial features, a well-known limitation referred to as spectral bias \cite{spectralbias, tancik2020fourier}, resulting in the generation of lower-quality frames. This is surprising, given that various position encoding methods---such as Fourier encoding \cite{mildenhall2020nerf, tancik2020fourier}---are well-established and widely used in tasks like image SR with INR due to their effectiveness, having become a conventional process \cite{lte, lee2022learning, btc, xiao2024towards}. 

Interestingly, however, we find that simply adding position encoding does not improve---and even degrades---performance in these models, an unexpected outcome that contrasts with the general success of position encoding in enhancing INR applications \cite{ijcai2021p151, gao2023adaptive, muller2022instant, kim2022scalable}. This issue becomes particularly pronounced when combined with pre-trained optical flow networks. We conjecture that, while these networks provide useful guidance for motion representation, integrating them with position encoding can inadvertently limit the model’s flexibility to fully leverage diverse video information.

To address these limitations, we propose BF-STVSR, a framework consisting of two modules: {\bspline} and {\fourier}, each designed to handle temporal and spatial features. First, {\bspline} utilizes B-spline basis functions, well-known established method for constructing smooth curves or surfaces \cite{btc}. This approach is well-suited for capturing the continuous nature of video motion. Next, {\fourier} represents spatial features by estimating dominant frequency information of input video frames, effectively capturing fine details. Additionally, unlike MoTIF \cite{chen2023motif}, {\bspline} models motion directly from encoded video features instead of relying on a pre-trained optical flow network. This not only allows the encoder to retain richer motion information for more accurate motion estimation but also improves efficiency by eliminating the need for the additional optical flow computation. Furthermore, our approach maintains reliable performance even without incorporating a pre-trained optical flow guidance in the training objective, further simplifying the overall framework.



In summary, our contributions are as follows: (1) We propose BF-STVSR, a framework consisting of two dedicated components, {\bspline} for temporal motion representation and {\fourier} for spatial feature representation, addressing the spatial and temporal axes independently. (2) BF-STVSR estimates motion directly from encoded video features, enhancing efficiency and simplifying the framework. (3) Our BF-STVSR achieves state-of-the-art performance on C-STVSR, demonstrating the effectiveness of our approach through extensive experiments.

\begin{figure*}
    \centering
    \includegraphics[width=\linewidth]{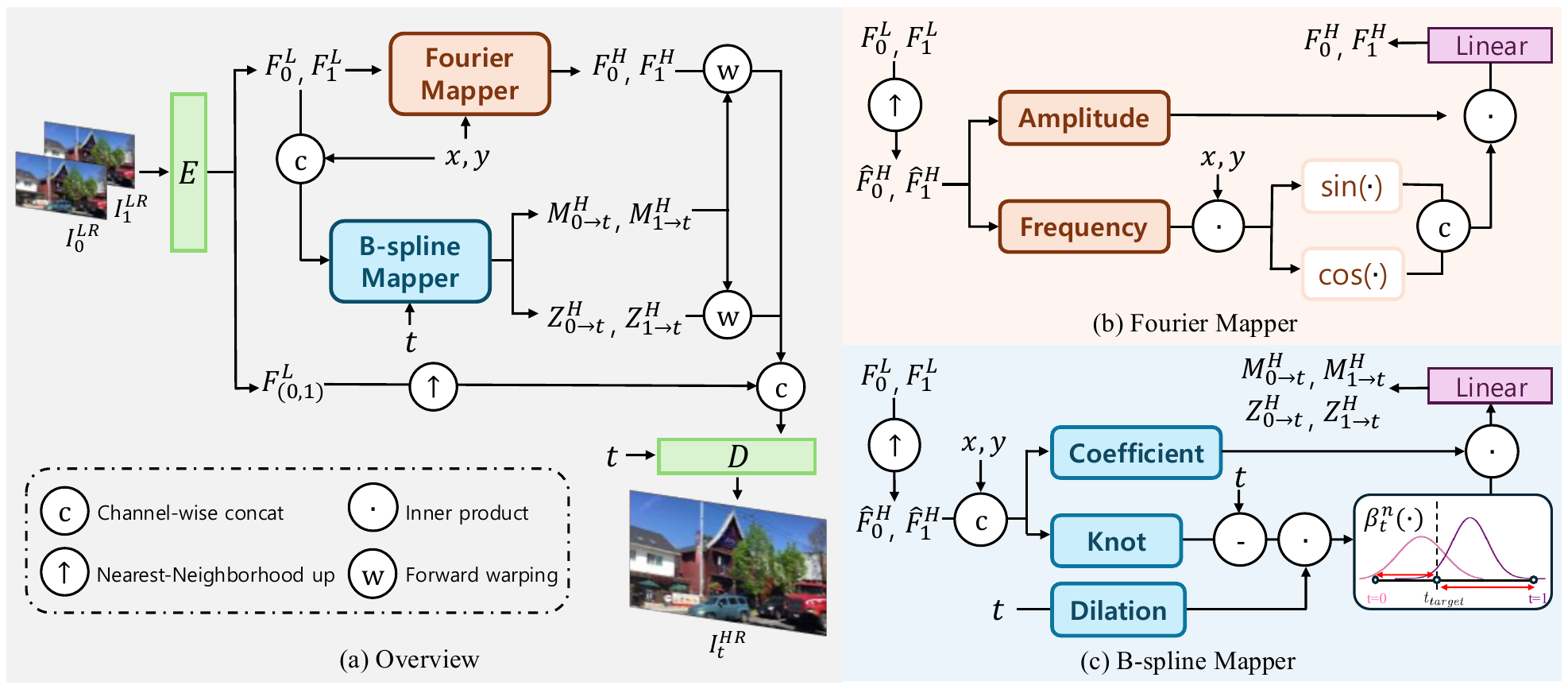} 

    \caption{\textbf{Schematic overview of our BF-STVSR.} (a) First, two input frames are encoded as low-resolution feature maps. Based on these features, {\fourier} predicts the dominant frequency information, while {\bspline} predicts smoothly interpolated motion representation, which is then processed into motion vectors at an arbitrary time $t$. The frequency information is temporally propagated by being warped with the predicted motion vectors. Finally, the warped feature is decoded to generate high-resolution interpolated RGB frame. (b) {\fourier} estimates the dominant frequencies and their amplitude to capture fine-detail information from the given frames. (c) {\bspline} estimates B-spline coefficients to model inherent motion, which smoothly interpolates motion features temporally.}
    \label{fig:main}
\end{figure*}

\section{Related Work}
\label{sec:formatting}

\subsection{Arbitrary Single Image Super-Resolution}

Single Image Super-Resolution (SISR) methods \cite{swinir, edsr, zhang2018image} have achieved impressive performance, but their reliance on fixed scales limits their applicability in real-world scenarios. To address this, several studies have proposed methods to perform super-resolution at arbitrary scales \cite{liif, ipe, lte, btc}. 
LIIF \cite{liif} introduced an Implicit Neural Representation (INR) for arbitrary scale image super-resolution, representing images continuously through local implicit functions.
IPE \cite{ipe} further used position encoding to address the spectral bias \cite{spectralbias}. Recently, LTE \cite{lte} proposed identifying dominant Fourier bases from latent features to effectively capture fine details and address spectral bias. Similarly, BTC \cite{btc} employed B-spline bases instead of Fourier bases to mitigate the Gibbs phenomenon observed in Screen Content Image Super-Resolution. Inspired by these methods, we explore effective position encoding techniques for C-STVSR, which reflect the characteristics of video data.

\subsection{Spatial-Temporal Video Super-Resolution}

While conventional Video Super-Resolution (VSR) \cite{caballero2017real, sajjadi2018frame,chan2021basicvsr, chan2022basicvsrpp} and Video Frame Interpolation (VFI) \cite{huang2022rife, niklaus2020softmax, park2021asymmetric, reda2022film, zhang2023extracting} 
perform interpolation along either spatial or temporal axis, Spatial-Temporal Video Super-Resolution (STVSR) conducts interpolation along both axes. Haris \textit{et al.} \cite{haris2020space} have introduced a unified framework for addressing STVSR and
Xiang \textit{et al.} \cite{xiang2020zooming} have proposed to use bidirectional deformable ConvLSTM. Although these studies demonstrate impressive performance in STVSR, they both have the limitation of only addressing STVSR at fixed scales. Recently, two works \cite{chen2022vinr, chen2023motif} have been proposed for Continuous Spatial-Temporal Video Super-Resolution (C-STVSR), which enables interpolation at arbitrary scales along both spatial and temporal axes. VideoINR \cite{chen2022vinr} is the first work on C-STVSR, which takes spatiotemporal coordinates as input and maps the corresponding RGB value in continuous manner using INR. Following this, MoTIF \cite{chen2023motif} generates temporal features using optical flows and performs forward warping to predict the interpolated high-resolution frame features. Although these studies effectively tackle the C-STVSR, relying solely on MLPs for spatial and temporal modeling leads to difficulties in learning the characteristics of the video. In this work, we adopt Fourier and B-spline basis functions to model spatial and temporal features of video data to address the aforementioned difficulties.

\section{Method}

\subsection{Overview} \label{3_overview}
The overall flow of our method, BF-STVSR, is illustrated in \Fref{fig:main} (a). Our framework is built upon the pipeline of MoTIF \cite{chen2023motif}, but differs in that it removes the need for an external optical flow network (e.g., RAFT~\cite{teed2020raft}) by introducing a learnable internal motion modeling approach based on B-spline and Fourier Mappers. Specifically, given two low-resolution frames $I_0^L, I_1^L \in \mathbb{R}^{3\times H \times W}$, our goal is to generate a high-resolution intermediate frame $I_t^H \in \mathbb{R}^{3\times sH \times sW}$ at any time $t \in [0, 1]$ with an arbitrary scale $s$. The encoder $E$ first takes the low-resolution frames as input and produces three latent features: $F_0^L, F_{(0,1)}^L, F_1^L \in \mathbb{R}^{C\times H \times W}$. Here, $F_0^L$ and $F_1^L$ represent the latent features of $I_0^L$ and $I_1^L$, while $F_{(0,1)}^L$ serves as a template feature for the intermediate frame, incorporating information from both input frames. The latent features $F_0^L$, $F_1^L$ are processed by (1) the {\bspline} (\Sref{3_B}), which predicts high-resolution motion vectors $M_{0 \rightarrow t}^H, M_{1 \rightarrow t}^H \in \mathbb{R}^{2\times sH \times sW}$ to the target time $t$, and (2) the {\fourier} (\Sref{3_F}), which estimates high-resolution spatial features $F_0^H$, $F_1^H \in \mathbb{R}^{C\times sH \times sW}$ at scale $s$. Finally, the high-resolution features $F_0^H, F_1^H$ are temporally propagated to the target time $t$ using forward warping based on the predicted motion vectors $M_{0 \rightarrow t}^H, M_{1 \rightarrow t}^H$, generating intermediate features $F_t^H$. These warped features are then concatenated with target time $t$ and $F_{(0,1)}^H$, a  nearest-neighbor upsampled $F_{(0,1)}^L$, and decoded to produce the high-resolution intermediate frame $I_t^H$.

\subsection{Temporal {\bspline}} \label{3_B}


Previous C-STVSR approaches~\cite{chen2022vinr, chen2023motif} employ implicit neural representations (INR) using MLPs that take spatiotemporal coordinates as input, enabling motion modeling at arbitrary target times $t$ and scales $s$. While INR-based motion modeling offers flexibility in motion prediction, we observe that it often struggles to effectively capture the complex and dynamic nature of motion in videos.

To better represent inherent motion, we introduce {\bspline}, which leverages the B-spline representation. B-spline bases are widely known for their effectiveness in modeling continuous signals \cite{btc}, making them well-suited for capturing smooth, continuous motion in videos, where objects move smoothly and continuously, rather than in jerky manner. The detailed process of {\bspline} is described in \Fref{fig:main} (b). We modify the Space-Time Local Implicit Neural Functions (ST-INF) from MoTIF \cite{chen2023motif}, resulting in our {\bspline}. Similar to ST-INF, {\bspline} predicts high-resolution forward motion vectors $M_{0 \rightarrow t}^H, M_{1 \rightarrow t}^H$ and reliability maps $Z_{0 \rightarrow t}^H, Z_{1 \rightarrow t}^H$ at arbitrary time $t \in [0, 1]$. A key difference is that our {\bspline} takes encoded features $F_0^L, F_1^L$ as input, rather than optical flows from an external network (e.g., RAFT \cite{teed2020raft}). 

In addition, rather than directly predicting motion vectors to the target time $t$, our {\bspline} $p_\psi$ models the inherent motion in the video by predicting B-spline coefficients and knots, as described in the following equation:
\begin{equation}
  p_\psi(z_r, \delta_r, \hat{t}) = c_r \odot \beta^n\left(\frac{\hat{t}-k_r}{d}\right).
  \label{eq:eq3}
\end{equation}
Here, $c_r=p_c(z_r,\delta_r)$, $k_r=p_k(z_r, \delta_r)$, and $d = p_d(g)$. Specifically, $z_r=F_{t_r}^L(q_r)$ is the latent feature vector at the coordinate $q_r = (x_r,y_r)$, nearest to the query coordinates $q = (x,y)$, with the reference frame time index $t_r \in \{0, 1\}$. The functions $p_c$, $p_k$, and $p_d$ are the estimators for the coefficients ($\mathbb{R}^{C+2} \mapsto \mathbb{R}^{C}$), knots ($\mathbb{R}^{C+2} \mapsto \mathbb{R}^{C}$), and dilation ($\mathbb{R}^{1} \mapsto \mathbb{R}^{C}$), respectively. $\hat{t} = |t - t_r|$ represents the relative temporal distance of the predicted feature to the reference frame, and $\delta_r(= q - q_r)$ is the spatial relative coordinate between the query and reference coordinates. Finally, $g$ is the frame interval of the input video.

After linearly projecting the predicted B-spline representation using $f_{\theta_b}$, we obtain the motion vector $M_{t_r \rightarrow t}^H(q)$ and reliability map $Z_{t_r \rightarrow t}^H(q)$ at the query coordinates $q$:
\begin{equation} \{Z_{t_r \rightarrow t}^H(q), M_{t_r \rightarrow t}^H(q)\} = f_{\theta_b} (p_\psi(z_r, \delta_r, \hat{t})). \label{eq:eq2
} \end{equation}
Using the predicted motion vectors, the spatial features $F^H_0$, $F^H_1$ and reliability maps are propagated to the target time $t$ via forward warping using softmax splatting \cite{niklaus2020softmax}. Finally, we obtain intermediate latent feature $F_t^H$ and corresponding reliability map $Z_t^H$. 
By directly learning the underlying motion from the input frames instead of individually predicting each arbitrary time $t$, our {\bspline} provides a more robust and flexible motion modeling approach. Note that, since our method does not rely on an external optical flow network, it offers more efficient and self-contained solution compared to prior approaches like MoTIF~\cite{chen2023motif}.

\begin{table*}[]
\caption{Performance comparison on the Fixed-scale STVSR baselines on Vid4, GoPro, and Adobe240 datasets. $\mathcal L_{RAFT}$ refers the optical flow supervision. Results are evaluated using PSNR (dB) and SSIM metrics. All frames are interpolated by a factor of $\times 4$ in the spatial axis and $\times 8$ in the temporal axis. ``\textit{Average}" refers to metrics calculated across all 8 interpolated frames, while ``\textit{Center}" refers to metrics measured using $1^{st}$, $4^{th}$ and $9^{th}$ (that is the center-frame interpolation) frames of the interpolated sequence. \textcolor{red}{Red} and \textcolor{blue}{blue} indicate the best and the second best performance, respectively.}
\label{tab:table_1}
\resizebox{\textwidth}{!}{%
\begin{tabular}{cc|cccccc}
\hline
\begin{tabular}[c]{@{}c@{}}VFI\\ Method\end{tabular} & \begin{tabular}[c]{@{}c@{}}VSR\\ Method\end{tabular} & Vid4          & GoPro-\textit{Center} & GoPro-\textit{Average} & Adobe-\textit{Center} & Adobe-\textit{Average} & \begin{tabular}[c]{@{}c@{}}Parameters\\ (Millions)\end{tabular} \\ \hline
SuperSloMo \cite{superslomo} & Bicubic & 22.42 / 0.5645 & 27.04 / 0.7937 & 26.06 / 0.7720 & 26.09 / 0.7435 & 25.29 / 0.7279 & 19.8 \\
SuperSloMo \cite{superslomo} & EDVR \cite{wang2019edvr} & 23.01 / 0.6136 & 28.24 / 0.8322 & 26.30 / 0.7960 & 27.25 / 0.7972 & 25.90 / 0.7682 & 19.8+20.7 \\
SuperSloMo \cite{superslomo} & BasicVSR \cite{chan2021basicvsr} & 23.17 / 0.6159 & 28.23 / 0.8308 & 26.36 / 0.7977 & 27.28 / 0.7961 & 25.94 / 0.7679 & 19.8+6.3 \\
\hline
QVI \cite{qvi_nips19} & Bicubic & 22.11 / 0.5498 & 26.50 / 0.7791 & 25.41 / 0.7554 & 25.57 / 0.7324 & 24.72 / 0.7114 & 29.2 \\
QVI \cite{qvi_nips19} & EDVR \cite{wang2019edvr} & 23.48 / 0.6547 & 28.60 / 0.8417 & 26.64 / 0.7977 & 27.45 / 0.8087 & 25.64 / 0.7590 & 29.2+20.7 \\
QVI \cite{qvi_nips19} & BasicVSR \cite{chan2021basicvsr} & 23.15 / 0.6428 & 28.55 / 0.8400 & 26.27 / 0.7955 & 26.43 / 0.7682 & 25.20 / 0.7421 & 29.2+6.3 \\
\hline
DAIN \cite{DAIN} & Bicubic & 22.57 / 0.5732 & 26.92 / 0.7911 & 26.11 / 0.7740 & 26.01 / 0.7461 & 25.40 / 0.7321 & 24.0 \\
DAIN \cite{DAIN} & EDVR \cite{wang2019edvr} & 23.48 / 0.6547 & 28.58 / 0.8417 & 26.64 / 0.7977 & 27.45 / 0.8087 & 25.64 / 0.7590 & 24.0+20.7 \\
DAIN \cite{DAIN} & BasicVSR \cite{chan2021basicvsr} & 23.43 / 0.6514 & 28.46 / 0.7966 & 26.43 / 0.7966 & 26.23 / 0.7725 & 25.23 / 0.7725 & 24.0+6.3 \\
\hline
\multicolumn{2}{c|}{ZoomingSloMo \cite{xiang2020zooming}} & 25.72 / 0.7717 & 30.69 / 0.8847 & - / - & 30.26 / 0.8821 & - / - & 11.10 \\

\multicolumn{2}{c|}{TMNet \cite{xu2021temporal}} & \textcolor{red}{25.96 / 0.7803} & 30.14 / 0.8696 & 28.83 / 0.8514 & 29.41 / 0.8524 & 28.30 / 0.8354 & 12.26 \\
\hline

\multicolumn{2}{c|}{VideoINR \cite{chen2022vinr}} & 25.61 / 0.7709 & 30.26 / 0.8792  & 29.41 / 0.8669 & 29.92 / 0.8746 & 29.27 / 0.8651 & 11.31 \\

\multicolumn{2}{c|}{MoTIF \cite{chen2023motif}}  & 25.79 / 0.7745 & 31.04 / 0.8877 & 30.04 / 0.8773 & 30.63 / 0.8839 & 29.82 / 0.8750 & 12.55 \\
\hline

\multicolumn{2}{c|}{BF-STVSR + $\mathcal L_{RAFT}$ (Ours)}  & 25.80 / 0.7754 & \textcolor{blue}{31.14 / 0.8893} & \textcolor{blue}{30.20 / 0.8799} & \textcolor{red}{30.84} / \textcolor{blue}{0.8877} & \textcolor{red}{30.14 / 0.8808} & \multirow{2}{*}{13.47} \\

\multicolumn{2}{c|}{BF-STVSR (Ours)}  & \textcolor{blue}{25.85 / 0.7772} & \textcolor{red}{31.17 / 0.8898} & \textcolor{red}{30.22 / 0.8802} & \textcolor{blue}{30.83} / \textcolor{red}{0.8880} & \textcolor{blue}{30.12} / \textcolor{red}{0.8808} &   \\

\hline
\end{tabular}%
}
\end{table*}

\begin{table*}[h]
\centering
\caption{Performance comparison on the C-STVSR baselines for out-of-distribution scale on GoPro dataset. $\mathcal L_{RAFT}$ refers the optical flow supervision. Results are evaluated using PSNR (dB) and SSIM metrics. All frames are interpolated by a scaling factor specified on the table and metrics calculated across all interpolated frames. \textcolor{red}{Red} and \textcolor{blue}{blue} indicate the best and the second best performance, respectively. }
\label{tab:table_2}
\resizebox{\textwidth}{!}{%
\begin{tabular}{c|c|cc|cc|c|c|c|c}
\hline 
\multirow{2}{*}{\shortstack{Temporal \\ Scale}} & \multirow{2}{*}{\shortstack{Spatial \\ Scale}} & \multicolumn{2}{c|}{RIFE \cite{huang2022rife}} & \multicolumn{2}{c|}{EMA-VFI \cite{zhang2023extracting}} & \multirow{2}{*}{\centering VideoINR \cite{chen2022vinr}} & \multirow{2}{*}{\centering MoTIF \cite{chen2023motif}} & \multirow{2}{*}{\shortstack{BF-STVSR \\ $+\mathcal L_{RAFT}$ (Ours)}} 
 & \multirow{2}{*}{\shortstack{BF-STVSR \\ (Ours)}} 
\\ 
\cline{3-6} 
& & \multicolumn{1}{c|}{LIIF \cite{liif}} & \multicolumn{1}{c|}{LTE \cite{lte}} & \multicolumn{1}{c|}{LIIF \cite{liif}} & \multicolumn{1}{c|}{LTE \cite{lte}} & & & \\ 
\hline \hline
$\times 8$ & $\times 4$ & \multicolumn{1}{c|}{29.14 / 0.8524}  & \multicolumn{1}{c|}{29.14 / 0.8524} & \multicolumn{1}{c|}{29.68 / 0.8671}  & \multicolumn{1}{c|}{29.68 / 0.8667} & 29.41 / 0.8669 & 30.04 / 0.8773 & \textcolor{blue}{30.20 / 0.8799} & \textcolor{red}{30.22 / 0.8802} \\ 
\hline
  & $\times 4$ & \multicolumn{1}{c|}{30.16 / 0.8738}  & \multicolumn{1}{c|}{30.16 / 0.8737} & \multicolumn{1}{c|}{30.64 / 0.8850}  & \multicolumn{1}{c|}{30.64 / 0.8848} & 30.78 / 0.8954 & 31.56 / 0.9064 & \textcolor{blue}{31.68 / 0.9082} & \textcolor{red}{31.70 / 0.9083} \\ 

$\times 6$ & $\times 6$ & \multicolumn{1}{c|}{27.87 / 0.8038}  & \multicolumn{1}{c|}{27.86 / 0.8031} & \multicolumn{1}{c|}{28.17 / 0.8126}  & \multicolumn{1}{c|}{28.17 / 0.8117} & 25.56 / 0.7671 & 29.36 / 0.8505 & \textcolor{blue}{29.44 / 0.8516} & \textcolor{red}{29.45 / 0.8520} \\ 

  & $\times 12$ & \multicolumn{1}{c|}{24.74 / 0.7019}  & \multicolumn{1}{c|}{24.70 / 0.6994} & \multicolumn{1}{c|}{24.85 / 0.7052}  & \multicolumn{1}{c|}{24.82 / 0.7028} & 24.02 / 0.6900 & \textcolor{red}{25.81 / 0.7330} & 25.78 / 0.7284 & \textcolor{blue}{25.80 / 0.7295} \\ 
\hline
  & $\times 4$ & \multicolumn{1}{c|}{27.43 / 0.8102}  & \multicolumn{1}{c|}{27.42 / 0.8100} & \multicolumn{1}{c|}{27.90 / 0.8263}  & \multicolumn{1}{c|}{27.90 / 0.8260} & 27.32 / 0.8141 & 27.77 / 0.8230 & \textcolor{blue}{28.06} / \textcolor{red}{0.8287} & \textcolor{red}{28.07 / 0.8287} \\ 

$\times 12$ & $\times 6$ & \multicolumn{1}{c|}{26.19 / 0.7640}  & \multicolumn{1}{c|}{26.19 / 0.7636} & \multicolumn{1}{c|}{26.49 / 0.7748}  & \multicolumn{1}{c|}{26.49 / 0.7743} & 24.68 / 0.7358 & 26.78 / 0.7908 & \textcolor{blue}{27.06 / 0.7961} & \textcolor{red}{27.07 / 0.7963} \\ 

   & $\times 12$ & \multicolumn{1}{c|}{24.03 / 0.6869}  & \multicolumn{1}{c|}{24.00 / 0.6853} & \multicolumn{1}{c|}{24.16 / 0.6918}  & \multicolumn{1}{c|}{24.15 / 0.6902} & 23.70 / 0.6830 & 24.72 / 0.7108 & \textcolor{blue}{24.87 / 0.7096} & \textcolor{red}{24.88 / 0.7104} \\ 
\hline

  & $\times 4$ & \multicolumn{1}{c|}{26.08 / 0.7735}  & \multicolumn{1}{c|}{26.08 / 0.7733} & \multicolumn{1}{c|}{\textcolor{red}{26.56 / 0.7904}}  & \multicolumn{1}{c|}{\textcolor{red}{26.56} / \textcolor{blue}{0.7902}}  & 25.81 / 0.7739 & 25.98 / 0.7758  & 26.40 / 0.7844 & 26.39 / 0.7840 \\ 

$\times 16$ & $\times 6$ & \multicolumn{1}{c|}{25.24 / 0.7394}  & \multicolumn{1}{c|}{25.24 / 0.7391} & \multicolumn{1}{c|}{25.54 / 0.7503}  & \multicolumn{1}{c|}{25.55 / 0.7499} & 23.86 / 0.7123 & 25.34 / 0.7527 & \textcolor{red}{25.81 / 0.7621} & \textcolor{red}{25.81} / \textcolor{blue}{0.7619} \\ 

  & $\times 12$ & \multicolumn{1}{c|}{23.57 / 0.6781}  & \multicolumn{1}{c|}{23.56 / 0.6769} & \multicolumn{1}{c|}{23.68 / 0.6828}  & \multicolumn{1}{c|}{23.69 / 0.6816} & 22.88 / 0.6659 & 23.88 / 0.6923 & \textcolor{red}{24.22} / \textcolor{blue}{0.6950} & \textcolor{red}{24.22 / 0.6955} \\ 
\hline
\end{tabular}%
}
\end{table*}



\subsection{Spatial {\fourier}} \label{3_F}

Even with the robust motion modeling provided by the {\bspline}, the quality of the interpolated feature $F_t^H$ depends significantly on the features propagated from $F_0^H$ and $F_1^H$. VideoINR \cite{chen2022vinr} and MoTIF \cite{chen2023motif} rely on simple MLPs to interpolate the latent features $F_0^L$ and $F_1^L$. However, implicit neural functions often struggle with capturing high-frequency details, leading to poor quality in the interpolated features, as noted in several studies \cite{spectralbias, tancik2020fourier, mildenhall2020nerf}. To address this issue, LTE \cite{lte} demonstrated that using Fourier bases for spatial feature modeling significantly improves performance in arbitrary-scale super-resolution by effectively capturing dominant frequencies. Inspired by this approach, we integrate a similar strategy into our Fourier Mapper. The detail process is illustrated in \Fref{fig:main} (c). The {\fourier} $g_\phi$ predicts the dominant frequencies and their amplitude of the Fourier bases for spatial features:
\begin{equation}
    \{F_0^H(q), F_1^H(q)\} = f_{\theta_f} (g_\phi(z_r, \delta_r)), \\
\end{equation}
\begin{equation}
    \text{where } g_\phi(z_r, \delta_r) \\
     = A_r \odot 
      \begin{bmatrix}
      \cos(\pi F_r \delta_r) \\
      \sin(\pi F_r \delta_r)
      \end{bmatrix}.
  \label{eq:eq4}
\end{equation}
Here, $A_r = g_a(z_r)$ and $F_r = g_f(z_r)$. Same as {\bspline}, $z_r = F_{t_r}^L(q_r)$ is the nearest latent feature vector from the query coordinates $q = (x, y)$ and $\delta_r(=q-q_r)$ is the relative coordinate in spatial domain. The $g_a$ and $g_f$ are the amplitude estimator ($\mathbb{R}^{C} \mapsto \mathbb{R}^{2C}$) and the frequency estimator ($\mathbb{R}^{C} \mapsto \mathbb{R}^{2C}$), respectively. By predicting dominant frequencies of query coordinates in latent space, {\fourier} improves the frequency details of the interpolated features $\hat{F}_0^H$ and $\hat{F}_1^H$. An additional linear projection $f_{\theta_f}$ is applied to the Fourier-embedded features, yielding refined representations of $F_0^H$ and $F_1^H$, which subsequently improve the quality of $F_t^H$. Although similar to LTE \cite{lte}, the proposed {\fourier} estimates amplitudes and frequencies from the nearest-neighbor interpolated $z_r$, and does not include a phase estimator.


\subsection{Training Objective} \label{3_Loss}



MoTIF \cite{chen2023motif} incorporates the optical flow supervision, resulting in the following training objective:
\begin{equation} \mathcal{L} = \mathcal{L}_{\text{char}}(\hat{I}_t^H, I_t^H) + \underbrace{\lambda \sum_{i=0}^{1} \mathcal{L}_{\text{char}}(\hat{M}_{i \rightarrow t}^H, M_{i \rightarrow t}^H)}_{\mathcal L_{RAFT}}, \label{eq:eq_flow_loss} \end{equation}
where $\mathcal{L}_{\text{char}}$ is the Charbonnier loss, $\hat{M}_{i \rightarrow t}^H$ and $M_{i \rightarrow t}^H$ are the RAFT-predicted and model-predicted motion vectors, respectively, $\hat{I}_t^H$ and $I_t^H$ are the ground-truth and predicted high-resolution frames at time $t$, and $\lambda$ is a hyperparameter. 

In contrast, our framework simplifies the objective by removing the optical flow supervision, $\mathcal L_{RAFT}$:
\begin{equation}
\mathcal{L} = \mathcal{L}_{\text{char}}(\hat{I}_t^H, I_t^H) 
\label{eq:eq_loss} 
\end{equation}
Despite this simplification, our model effectively estimates motion, achieving performance comparable to, or even better than, models trained with the optical flow supervision, $\mathcal L_{RAFT}$.




\begin{figure*}[!t]
    \centerline{\includegraphics[width=\linewidth]{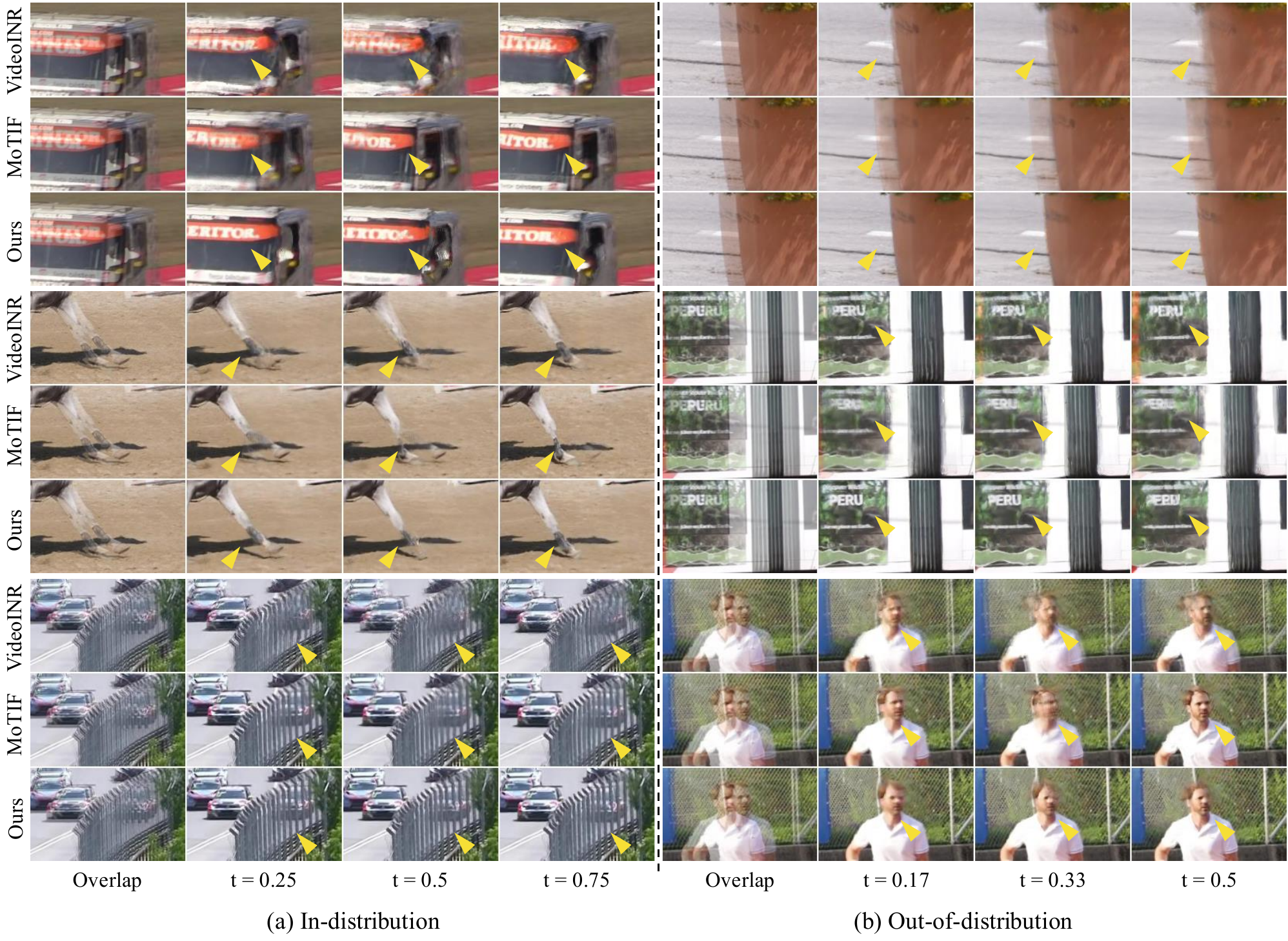}}
    \caption{Qualitative comparison on arbitrary scale temporal interpolation. ``Overlap'' refers to the averaged image of two input frames ($t=0,1$), and the following images are interpolated results at $t \in [0,1]$. (a) shows the interpolated results on in-distribution temporal scale ($\times 8$), used during training. (b) shows the interpolated results on out-of-distribution temporal scale ($\times 6$), not seen during training. }
    
    \label{figure3} 
\end{figure*}

\begin{table*}[]
\centering
\caption{Performance comparison on the one-stage C-STVSR baselines on GoPro and Adobe240 datasets. $\mathcal{L}_{RAFT}$ refers to the optical
flow supervision. Results are evaluated using VFIPS \cite{hou2022vfips}, FloLPIPS \cite{danier2022flolpips}, tOF \cite{chu2020tecoGAN}, and VMAF \cite{li2016vmaf} metrics. All frames are interpolated by a factor of $\times 4$ in the spatial axis and $\times 8$ in the temporal axis. \textcolor{red}{Red} and \textcolor{blue}{blue} indicate the best and the second best performance, respectively.}
\label{tab:vid_metric}
\resizebox{0.8\textwidth}{!}{
\begin{tabular}{@{}cccccccccc@{}}
\toprule
\multirow{2}{*}{Method} & \multicolumn{4}{c}{GoPro} & \multicolumn{4}{c}{Adobe} \\ 
\cmidrule(lr){2-5} \cmidrule(lr){6-9}
& VFIPS$\uparrow$ & FloLPIPS$\downarrow$ & tOF$\downarrow$ & VMAF$\uparrow$ & VFIPS$\uparrow$ & FloLPIPS$\downarrow$ & tOF$\downarrow$ & VMAF$\uparrow$ \\ 
\midrule
VideoINR  & 81.13 & \textcolor{red}{0.151} & 0.519 & 57.96 & 81.15 & 0.145 & 0.574 & 67.08 \\
MoTIF     & 81.89 & \textcolor{blue}{0.156} & 0.517 & 59.82 & 81.61 & 0.144 & 0.607 & 68.40 \\
BF-STVSR + $\mathcal{L}_{RAFT}$ (Ours) & \textcolor{red}{83.26} & \textcolor{red}{0.151} & \textcolor{red}{0.474} & \textcolor{red}{61.09} & \textcolor{red}{84.14} & \textcolor{red}{0.131} & \textcolor{red}{0.488} & \textcolor{blue}{70.79}  \\
BF-STVSR (Ours) & \textcolor{blue}{83.01} & \textcolor{red}{0.151} & \textcolor{blue}{0.480} & \textcolor{blue}{61.06} & \textcolor{blue}{84.04} & \textcolor{blue}{0.132} & \textcolor{blue}{0.498} & \textcolor{red}{70.82}  \\
\bottomrule
\end{tabular}
}
\end{table*}


\section{Experiments}
\label{sec:experiments}
\subsection{Experiments Setup}

\paragraph{Implementation and Training Details}

We follow the same training scheme as \cite{chen2022vinr, chen2023motif} unless otherwise noted. We adopt the same two-stage training strategy: for the first 450,000 iterations, the spatial scaling factor is fixed as 4, while for the remaining 150,000 iterations, it is uniformly sampled from $[2,4]$. The $\lambda$ is set as 0.01. We use the Adam optimizer with parameters $\beta_1 = 0.9$ and $\beta_2 = 0.999$, and apply cosine annealing to decay the learning rate from $10^{-4}$ to $10^{-7}$ for every 150,000 iterations. ZoomingSlowMo \cite{xiang2020zooming} is used as the encoder, with a batch size of 32, and random rotation and horizontal-flipping for data augmentation. To ensure training stability, we substitute the predicted forward motion with the ground-truth forward motion with a certain probability, starting from 1.0 and gradually reducing to 0 over the first 150,000 iterations. For {\bspline}, we use the three-layer SIRENs \cite{siren} as the coefficient and knot estimators, and a single fully connected layer as the dilation estimator. In {\fourier}, we use three-layer SIRENs as the amplitude and frequency estimators, followed by a 3$\times$3 convolutional layer for spatial encoding. Both {\bspline} and {\fourier} have hidden dimensions of 64, with SIREN layer dimensions set to 64, 64, and 256. 

\paragraph{Datasets}
We use the Adobe240 dataset \cite{su2017deep} for training, which consists of 133 videos in 720P taken by hand-held cameras. During training, nine sequential frames are selected from the video and the $1^{st}$ and $9^{th}$ frames are used as input reference frames. Three frames are then randomly sampled between them and used as the target ground-truth frames. 
For evaluation, we use Vid4\cite{vid4}, Adobe240 \cite{adobe240}, and GoPro \cite{Nah_2017_CVPR} datasets. Unless otherwise specified, the default spatial scale is 4. For Vid4, temporal scale is set to $\times$2, corresponding to the center-frame interpolation. For Adobe240-\textit{Average} and GoPro-\textit{Average}, the temporal scale is set as $\times$8, representing multi-frame interpolation. Additionally, for Adobe240-\textit{center} and GoPro-\textit{Center}, evaluation is performed only on $1^{st}$, $4^{th}$, $9^{th}$ frames, representing the center-frame interpolation.

\paragraph{Baseline methods}
We categorize baseline models into two types---continuous and fixed-scale---and conduct comparisons within each category. Here, Fixed-scale Spatial-Temporal Video Super-Resolution (Fixed-STVSR) are limited to super-resolving at fixed scaling factors in both axes that are learned during the training. First, we select two-stage Fixed-STVSR methods that combine fixed video super-resolution models (\textit{e.g.}, Bicubic Interpolation, EDVR \cite{wang2019edvr}, BasicVSR \cite{chan2021basicvsr}) with video frame interpolation models (\textit{e.g.}, SuperSloMo \cite{superslomo}, QVI \cite{qvi_nips19}, DAIN \cite{DAIN}). Second, we select one-stage Fixed-STVSR method, specifically ZoomingSlowMo \cite{xiang2020zooming}. 
For continuous methods, we select two-stage C-STVSR methods that combine continuous image super-resolution models (\textit{e.g.}, LIIF \cite{liif}, LTE \cite{lte}) with video frame interpolation models (\textit{e.g.}, RIFE \cite{huang2022rife}, EMA-VFI \cite{zhang2023extracting}). Lastly, we select one-stage C-STVSR methods, including TMNet \cite{xu2021temporal}, which is limited to $\times$4 spatial super-resolution, VideoINR \cite{chen2022vinr}, and MoTIF \cite{chen2023motif}.


\paragraph{Evaluation Metrics}
We evaluate model performance using PSNR and SSIM on the Y channel. To assess video quality, we employ VFIPS \cite{hou2022vfips} and FloLPIPS \cite{danier2022flolpips} that primarily designed for VFI to capture perceptual similarity. Additionally, we report tOF \cite{chu2020tecoGAN} to measure temporal consistency based on the optical flow. To further evaluate video quality, we utilize VMAF \cite{li2016vmaf}, a perceptual metric developed for real-world video streaming applications. We measure the average VMAF score for videos encoded at 30 fps.

\subsection{Quantitative results}
\begin{figure}[th!]
  \centering
    \includegraphics[width=\linewidth]{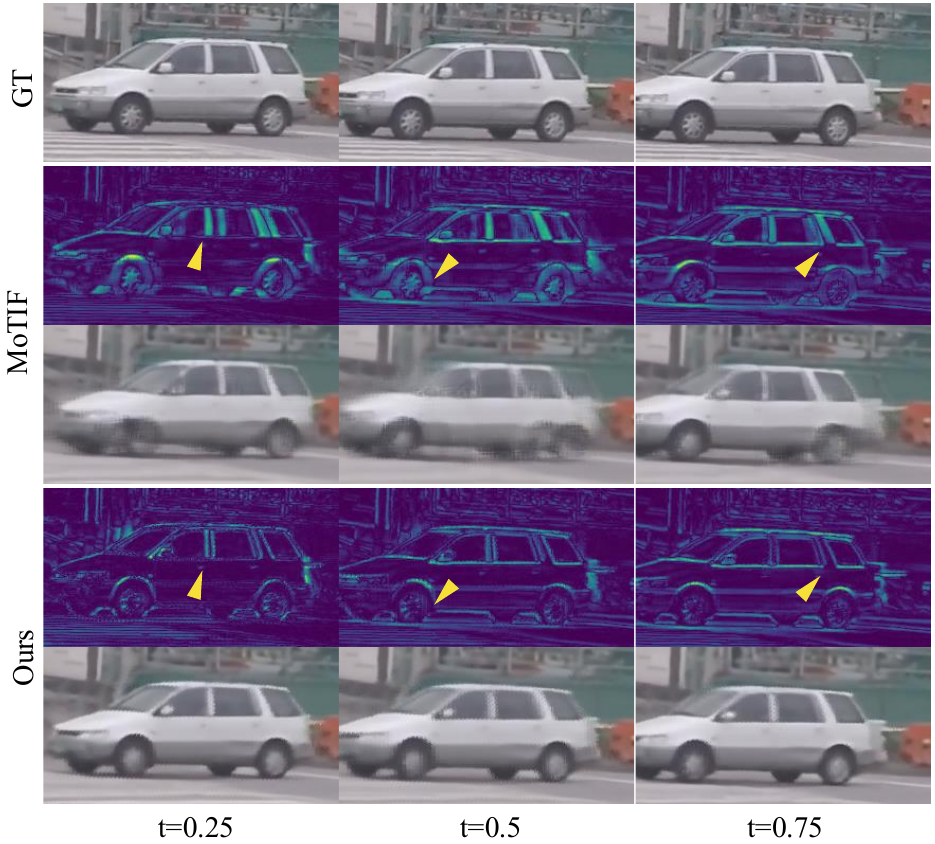}

   \caption{Qualitative comparison on the large out-of-distribution scale with a spatial scale of $\times\text{4}$ and a temporal scale of $\times\text{12}$. Three interpolation results at $t= \text{0.25, 0.5, 0.75}$ are shown with residual intensity maps compared to the ground truth frames.}
   \label{fig:fig4}
\end{figure}


We compare our model with Fixed-STVSR methods in \Tref{tab:table_1}. For center-frame interpolation tasks in STVSR, including Vid4, GoPro-\textit{Center}, and Adobe-\textit{Center}, our model achieves the best performance on all datasets except Vid4. On Vid4, TMNet outperforms other models, likely due to its training on Vimeo90K dataset \cite{xue2019video}, which shares similar characteristics with Vid4. For multi-frame interpolation tasks in STVSR, represented by GoPro-\textit{Average} and Adobe-\textit{Average}, our model surpasses the performance of the state-of-the-art MoTIF, which uses a pre-trained optical flow network \cite{teed2020raft} to generate temporal features during training. This improvement suggests that the {\bspline} and {\fourier} provide more robust temporal and spatial feature representations. We also evaluate our model against one-stage C-STVSR methods using video quality metrics in \Tref{tab:vid_metric}. Our model consistently outperforms the baselines across all metrics by a significant margin, except for the FloLPIPS on GoPro dataset. This demonstrates the superior temporal consistency and perceptual quality of the proposed method. \Tref{tab:table_2} compares the performance of the proposed method with C-STVSR methods for out-of-distribution scales on GoPro dataset. BF-STVSR achieves the best performance across all test cases, except at a $\times$16 temporal scale and $\times$4 spatial scale. This suggests that our {\bspline} generalizes better to unseen time intervals and effectively handles temporal interpolation. Note that in all test cases, our model performs comparably to the one with $\mathcal{L}_{RAFT}$.

\subsection{Qualitative results}
\Fref{figure3} presents qualitative results comparing our model with VideoINR and MoTIF. The results include interpolated frames for an in-distribution temporal scale ($\times$8), used during training (left), and an out-of-distribution temporal scale ($\times$6), unseen during training (right). For the in-distribution scale, BF-STVSR captures high-frequency details more effectively, particularly in the horse’s hooves and the striped shape of the handrails. For the out-of-distribution scale, BF-STVSR demonstrates superior performance in dynamic motion scenes, accurately interpolating edges of the text and the man’s face, where other methods produce blurry or ghosted frames. These results highlight our model’s ability to perform natural motion interpolation for moving objects while effectively preserving high-frequency details. 
Additionally, \Fref{fig:fig4} shows interpolated results at an extreme scale with a spatial scale of $\times$4 and a temporal scale of $\times$12. We include interpolated frames at sampled time points ($t = 0.25, 0.5, 0.75$) along with residual intensity maps compared to ground truth frames. Our method produces sharper and more accurate results than MoTIF, especially in areas like the tire and the region next to the car window. 
\subsection{Computational Cost and Latency}
\begin{figure}[h]
  \centering
    \includegraphics[width=\linewidth]{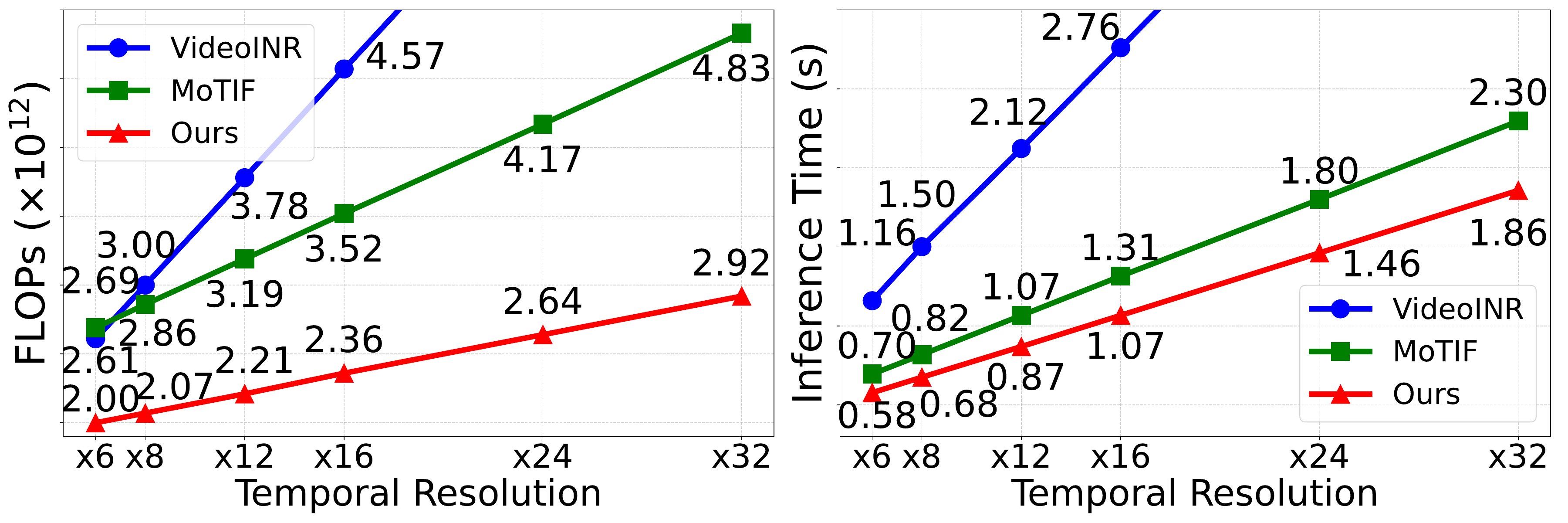}
   \caption{Computational cost (left) and inference time (right) comparison on the spatial resolution of $1280\times 720$ with different temporal scale. All frames are spatially interpolated by a factor of ×4.}
   \label{fig:inferencetime}
\end{figure}

\noindent To evaluate the computational efficiency of our method, we compare the FLOPs and inference time of the baselines \cite{chen2022vinr, chen2023motif} and our method across different temporal scales in \Fref{fig:inferencetime}. We use the \textbf{fvcore} library\footnote{https://github.com/facebookresearch/fvcore} to measure FLOPs and benchmark average inference time over 100 iterations on an NVIDIA RTX 4090 GPU. The evaluation is conducted on the spatial resolution of $1280\times 720$, with a spatial upscaling factor of $\times4$ and varying temporal scales. To efficiently evaluate the B-spline function, we implement a CUDA kernel. Our method removes additional optical flow computations, enhancing efficiency. Once predicted, the B-spline representation enables lightweight motion estimation at each time step through simple linear projection, further reducing computational overhead. As shown in \Fref{fig:inferencetime}, our method consistently achieves the lowest computational cost and fastest inference across all temporal resolutions. 

\subsection{Optical Flow and Position Embeddings}

\begin{table}[h]
\centering
\setlength{\tabcolsep}{9pt}
\caption{The impact of different position embeddings and the pre-trained optical flow network. \textbf{O{$\cdot$}F} denotes using pre-trained RAFT \cite{teed2020raft} for motion modeling, $\mathcal L_{RAFT}$ refers the optical flow supervision, \textbf{B} represents {\bspline} and \textbf{F} represents {\fourier}. The first row corresponds to the default MoTIF \cite{chen2023motif}. Results are evaluated using PSNR (dB) and SSIM metrics.}

\label{tab:optical_abl}
\resizebox{\linewidth}{!}{%
\begin{tabular}{cccc||cc}
\hline
\textbf{O{$\cdot$}F} & \textbf{B} & \textbf{F} & \textbf{$\mathcal L_{RAFT}$} & GoPro-\textit{Average}  & Adobe-\textit{Average}  \\ 
\hline 
\hline
{\cmark}   &  & & {\cmark} & 30.04 / 0.8773 & 29.82 / 0.8750 \\ 
\hline
{\cmark} &  & {\cmark}  & {\cmark} & 29.94 / 0.8764 & 29.73 / 0.8741 \\ 
\hline
{\cmark}  & {\cmark}  &   & {\cmark} & 30.03 / 0.8774 & 29.81 / 0.8756 \\ 
\hline 
\hline
 &  &  {\cmark} & {\cmark} & 30.12 / 0.8783 & 30.02 / 0.8784 \\ 
\hline
 & {\cmark} &  & {\cmark} & 30.16 / 0.8792 & 30.11 / 0.8801 \\ 
\hline
 & {\cmark} & {\cmark} & {\cmark} & 30.20 / 0.8799 & \textcolor{red}{30.14 / 0.8808} \\ 
\hline
 & {\cmark} & {\cmark} &  & \textcolor{red}{30.22 / 0.8802} & 30.12 / \textcolor{red}{0.8808} \\ 
\hline
\end{tabular}%
}
\end{table}

\noindent\Tref{tab:optical_abl} compares model performance with and without the pre-trained optical flow network, RAFT \cite{teed2020raft}, for motion modeling and the optical flow supervision, $\mathcal L_{RAFT}$, across different combinations of our proposed {\bspline} and {\fourier}. The first row shows the basic MoTIF \cite{chen2023motif} configuration. As seen in the second and third row, including the optical flow network with the proposed modules degrades performance. In contrast, directly using the proposed modules to extract spatial and temporal features, without the optical flow network, improves performance across all cases (last four rows). Note that even without $\mathcal L_{RAFT}$, our proposed model achieves similar or better performance (last row). We attribute this improvement to the ability of the proposed modules to effectively extract and utilize the rich information embedded within the video, thereby enhancing the model’s capacity to capture complex spatial and temporal features. 
Additionally, as shown in the fourth and fifth rows of the table, performance decreases when each mapper is used independently, but the best results are achieved when both mappers are integrated.

\begin{figure}[h]
  \centering
    \includegraphics[width=0.9\linewidth]{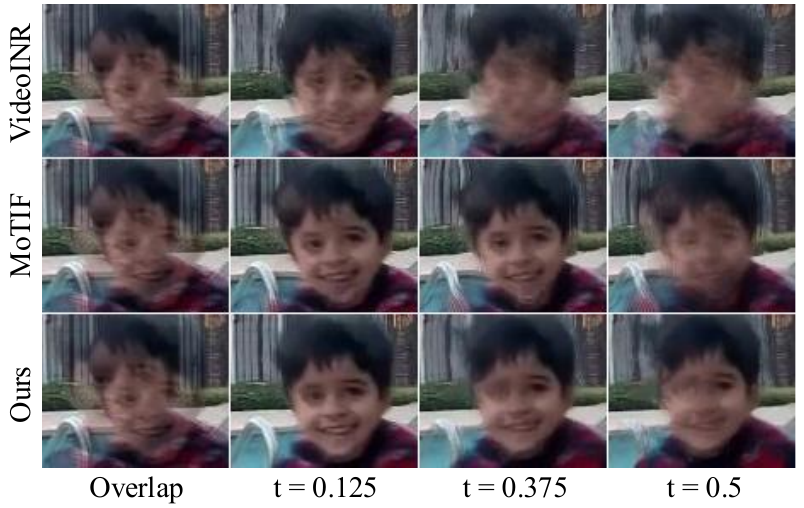}
   \caption{Qualitative comparison on a large motion case with a spatial scale of $\times 1$ and a temporal scale of $\times 8$. Three interpolation results at $t= \text{0.125, 0.375, 0.5}$ are shown.}
   \label{fig:limitations}
\end{figure}



\paragraph{Limitations} While our method demonstrates performance improvements, there still remain certain limitations. As shown in \Fref{fig:limitations}, existing C-STVSR models, including ours, still struggle with handling large motion. Moreover, the training process of C-STVSR models is time-consuming and computationally expensive. Addressing these challenges is left for future work.

\section{Conclusions}

In this paper, we proposed BF-STVSR, a novel framework for Continuous Spatial-Temporal Video Super-Resolution (C-STVSR). Motivated by our observation that na\"{i}ve position encoding can degrade performance---particularly when paired with optical flow networks---we introduced two axis-specific position encoding modules: {\bspline}, which leverages B-spline basis functions for smooth and accurate temporal interpolation, and {\fourier}, which captures dominant spatial frequencies to effectively model fine-grained spatial details. 
By estimating motion directly from encoded features, our design eliminates the need for external optical flow supervision, achieving high efficiency while maintaining strong performance. Extensive experiments confirm that BF-STVSR achieves state-of-the-art results in PSNR, SSIM and various video quality metrics, demonstrating superior spatial detail, natural temporal consistency, and robustness under challenging conditions, including extreme out-of-distribution scales.

\clearpage
{
    \small
    \paragraph{Acknowledgement}
This work was supported by the National Research Foundation of Korea (NRF) grant funded by the Korea government (MSIT) (No.2022R1C1C100849612) and Institute of Information \& communications Technology Planning \& Evaluation (IITP) grant funded by the Korea government (MSIT) (No.RS-2020-II201336, Artificial Intelligence Graduate School Program (UNIST), No.2022-0-00959, No.RS-2022-II220959 (Part 2) Few-Shot Learning of Causal Inference in Vision and Language for Decision Making, RS-2022-II220264, Comprehensive Video Understanding and Generation with Knowledge-based Deep Logic Neural Network), and Artificial intelligence industrial convergence cluster development project funded by the Ministry of Science and ICT(MSIT, Korea) \& Gwangju Metropolitan City. This research used high performance computing resources of the UNIST Supercomputing Center.

    \bibliographystyle{ieeenat_fullname}
    \bibliography{main}
}




\end{document}